\documentclass[10pt,journal,compsoc]{IEEEtran}
\ifCLASSOPTIONcompsoc
  \usepackage[nocompress]{cite}
\else
  \usepackage{cite}
\fi
\usepackage{graphicx}
\usepackage{subfigure}
\usepackage[cmex10]{amsmath}
\usepackage{mdwmath}
\usepackage{mdwtab}
\usepackage{booktabs}

\ifCLASSINFOpdf
\else
\fi
\begin{document}
%
\title{Multiple Domain Cyberspace Attack and Defense Game Based on Reward Randomization Reinforcement Learning}
%
%
%
%

\author{Lei~Zhang,
        Yu~Pan,
        Yi~Liu,
        Qibin~Zheng
        and~Zhisong~Pan 
\IEEEcompsocitemizethanks{\IEEEcompsocthanksitem Corresponding author: Zhisong Pan. E-mail:panzhisong@aeu.edu.cn.
\IEEEcompsocthanksitem Both Lei Zhang and Yu Pan are contributed equally to this research.
\IEEEcompsocthanksitem This work was supported by the National Natural Science Foundation of China No.62076251 and No.62106281.
\IEEEcompsocthanksitem Lei Zhang, Yu Pan and Zhisong Pan are with Command and Control Engineering College, Army Engineering University of PLA, Nanjing, Jiangsu, China, 210007. (E-mail: zhanglei@aeu.edu.cn, panyu@aeu.edu.cn, panzhisong@aeu.edu.cn).\protect
\IEEEcompsocthanksitem Yu Pan is with 31436 troop of PLA, Shenyang, Liaoning, China, 110005. (E-mail: panyu@aeu.edu.cn).
\IEEEcompsocthanksitem Yi Liu and Qibin Zheng are with Academy of Military Science, Haidian, Beijing, China, 100091. (E-mail: albertliu20th@163.com, zqb1990@hotmail.com).}
\thanks{Manuscript received xxxx, 2022; revised xxxx, 2022.}}

%
%

\markboth{IEEE Transcations on \LaTeX\ Class Files,~Vol.~16, No.~7, xxxx~2022}%
{Shell \MakeLowercase{\textit{et al.}}: Bare Demo of IEEEtran.cls for Computer Society Journals}
%



\IEEEtitleabstractindextext{%
\begin{abstract}
The existing network attack and defense method can be regarded as game, but most of the game only involves network domain, not multiple domain cyberspace. To address this challenge, this paper proposed a multiple domain cyberspace attack and defense game model based on reinforcement learning. We define the multiple domain cyberspace include physical domain, network domain and digital domain. By establishing two agents, representing the attacker and the defender respectively, defender will select the multiple domain actions in the multiple domain cyberspace to obtain defender's optimal reward by reinforcement learning. In order to improve the defense ability of defender, a game model based on reward randomization reinforcement learning is proposed. When the defender takes the multiple domain defense action, the reward is randomly given and subject to linear distribution, so as to find the better defense policy and improve defense success rate. The experimental results show that the game model can effectively simulate the attack and defense state of multiple domain cyberspace, and the proposed method has a higher defense success rate than DDPG and DQN.
\end{abstract}

}

\maketitle

\IEEEdisplaynontitleabstractindextext

%
\IEEEpeerreviewmaketitle

\IEEEraisesectionheading{\section{Introduction}\label{sec:introduction}}

%
%
%
%

\IEEEPARstart{W}{ith} the continuous development of the information society, cyberspace attacks are becoming more and more frequent. In order to improve the overall security of the cyberspace, we study and explore the cyberspace security defense technology system from the perspective of multiple domain cyberspace attack and defense confrontation, and improve the cyberspace security defense ability, which has important theoretical value and practical significance. In reality, we need to fully and effectively analyze cyberspace attacks, so that administrator can use limited cyberspace resources and devices to make better cyberspace security defense policies, so as to improve the effectiveness of cyberspace defense. At the same time, actively building an intelligent defense system for cyberspace security has become a new research direction and hotspot to solve cyberspace security problems in recent years. Meanwhile, game theory is highly consistent with the antagonism of goals, non-cooperation and strategy dependence of both sides of the cyberspace attack and defense. Therefore, the defense method based on game model has become one of the important methods of cyberspace security defense.

By strengthening the learning and training of an agent, the agent can have a strong ability to observe the cyberspace state and master the rules of the attackers, so that the agent can capture or expel the attackers, improve the overall security of cyberspace, and reduce the cost of cyberspace management. For administrators, however, his management action is more than simple view of the internal information is being attacked by the attacker, and should have to modify the firewall configuration, cut traffic access, restart the service and so on. Therefore, in order to better simulate multiple domain cyberspace game environment, we proposed the multiple domain cyberspace attack and defense game model based on reinforcement learning. 

Cyberspace attack and defense is essentially a process of adversary. The success of cyberspace depends not only on the strength of the attacker's own attack ability, but also on the targeted defense actions taken by the defender against the attacker. Therefore, the process of cyberspace attack and defense is a process of adversary. Since Nash initiated game theory, game theory has developed into a systematic discipline theory, and has been widely used in various fields, especially in the modern basic theory of economics. Game theory analysis has played an important role.

Although the basic theory of game theory and related game models are proposed based on various problems existing in economics, the cooperative versus competitive mechanism described by game models also exists widely in all aspects, so it is increasingly applied to other fields. This paper focuses on how to apply the game theory to the field of cyberspace attack and defense, improve the defense ability of defenders, and make the cyberspace defense system more efficient and practical. Cyberspace attack and defense game model is of great significance to the study of cyberspace attack and defense technology in the era of artificial intelligence. Firstly, the game model shifts the research focus from the specific attacking action of the attacker to the adversary system composed of the attacker and the defender. Secondly, the game model includes the key factors of the cyberspace attack and defense adversary process, such as incentive, utility, cost, risk, policy, action, security mechanism, security measure, attack means, protection means, system state, etc. Finally, the equilibrium strategy of cyberspace attack and defense can be deduced by using the cyberspace attack and defense game model \cite{gottlob:nonmon}.

The main contributions of this paper are as follows:
\begin{itemize}
	\item{We propose an attacker and defender game model in multiple domain cyberspace. In general, the existing attacker and defender game model is based on network domain, it doesn't take into account the multiple domain;}
	
	\item{We proposed a reward randomization reinforcement learning that for discovering more game optimal policies. Experimental results show that reward randomization DDPG can find more attackers, can counter different attack policies;}
	
\end{itemize}

The rest of this paper is organized as follows. In the next section we introduce the related works. An attacker and defender game model is proposed in Section 3. Experiments and discussions in Section 4. Finally, we summarize the paper in Section 5.

\section{Related Works}

Existing cyberspace attack and defense based on game theory has made relevant results in the study of attack and defense game, but the current research is basically a perfectly rational assumptions, and the type of game can be divided into single stage game and multiple stage game. Static game theory is used to attack worm virus, and better results are obtained \cite{gottlob:noon}. Also, other researcher constructs a non-cooperative game model between the attacker and the sensor trust node, and obtains the optimal attack policy according to Nash equilibrium \cite{gottlob:noonnnn}. However, in general cyberspace attack and defense game, the application of single stage game is small. Most cyberspace attack and defense game should be in multiple stage game, and the attack and defense process usually lasts several stages, which is similar to the multiple domain cyberspace attack and defense policy. A multiple stage game model based on the cyberspace attack and defense game was proposed and the solution method of Nash equilibrium was further given \cite{Sarta2020DynamicSG}. In this model, the administrator was taken as the source of signal transmission and the attacker was taken as the receiver of signal. Also, the multiple stage game model of intrusion detection system and wireless sensor was established \cite{Chowdhary2019AdaptiveMS}. The model of attack tree is a formal modeling method to describe system security \cite{Lallie2020ARO}. It represents an attack against the system as a tree structure, taking the target to be reached by the attack as the root node of the tree, and taking the different methods to achieve the target as the leaf node. The above research analyzed the multiple stage game model, but the system environment was affected by various aspects and had certain randomness.

Random game is a multiple stage game model, which uses Markov process to describe the state transition, so it can analyze the influence of randomness on the cyberspace attack and defense game process. The cyberspace attack and defense problem can directly abstracted into random game problem \cite{Liu2020OptimalND}, and the benefits of relevant attack and defense game are given. Following, the convex analysis is used to transform the Nash equilibrium problem into a nonlinear programming problem, so as to solve the nonlinear programming problem. A game model between malware and security software is proposed, based on which a game model is constructed for defense . Most of the research mentioned above is based on completely rational assumption and does not consider the multiple domain cyberspace. Therefore, it has important research value to study the multiple domain cyberspace attack and defense game. However, there is too much information exchange between participants in game, and there is not enough research on strategy adjustment and stability of attacker.



Reinforcement learning is a classical machine learning method. It mainly learns through the interaction between an agent and the environment. Compared with other machine learning methods, reinforcement learning is more suitable for independently guiding the policy decisions of both sides of the game. In this paper, the reinforcement learning is introduced into the game, the game method is introduced into the multiple domain cyberspace attack and defense, and the game method is adopted to study the cyberspace attack and defense state, which is more in line with the realistic cyberspace environment. At the same time, the reinforcement learning is more suitable for guiding the decision making of the attack and defense agents.

\textbf{Deep Deterministic Policy Gradient (DDPG)}: DDPG \cite{Continuous} is a learning method that integrates deep learning neural network into Deterministic Policy Gradient (DPG) \cite{Deterministic}. Compared with DPG, the improvement the use of neural network as policy network and \emph{Q}-network, then used deep learning to train the above neural network. DDPG has four networks: actor current network, actor target network, critic current network and critic target network. In addition to the four network, DDPG also uses experience playback, which is used to calculate the target \emph{Q}-value. In DQN, we are copying the parameters of the current \emph{Q}-network directly to the target \emph{Q}-network, that is $\theta^{Q '}=\theta^Q$, but DDPG use the following update:

\begin{equation}\label{equation1}
	\left  \{
	\begin{array}{l}
		{\theta ^{Q'}} \leftarrow \tau {\theta ^Q} + (1 - \tau ){\theta ^{Q'}}\\\\
		{\theta ^{\mu '}} \leftarrow \tau {\theta ^\mu } + (1 - \tau ){\theta ^{\mu '}}
	\end{array}
	\right.
\end{equation}

where $\tau$ is the update coefficient, which is usually set as a small value, such as 0.1 or 0.01. And this is the loss function:

\begin{equation}
	L(w) =\frac{1}{m}\sum\limits_{j=1}^m(y_j-Q(\phi(S_j),A_j,w))^2
\end{equation}

\section{Attack and Defense Game Model}

\subsection{Definition of Multiple Domain Cyberspace}

With the deepening of the understanding of the concept of cyberspace, especially the concept of cyberspace, more and more researchers realize that cyberspace is affected by multiple domain action. Cyberspace can be thought of as merging in multiple domains such as the physical domain, the digital domain, the network domain and the social domain, underlining the multiple domain feature of cyberspace. In this paper, because the social domain mainly involves the social relationship between the defender (administrator) and the attacker, this paper does not discuss the social domain feature of multiple domain cyberspace, we define the multiple domain cyberspace includes physical domain, digital domain and network domain. The physical domain describes device information in the space, like room, building, computer terminal, server terminal. In physical domain, attacker and defender have enter room, out room, control computer terminal and other physical domain actions. The network domain describes interface, path, and action device information related to network transmission, that is impressions of the general cyberspace for most of people. At last, the digital domain describes digital information in cyberspace, like username, password, etc.

In addition, we have some network security protection rules in cyberspace, including physical domain security protection rules, network domain security protection rules and digital domain security protection rules. Physical domain security protection rules mainly describe how to prevent unauthorized access. Physical domain, which prevents an attacker from entering a specific space. Network domain security protection rules mainly describe the methods to prevent unauthorized access in network domain. Generally, access control lists, static routes, and VLAN partitions can be used to achieve network isolation. In this paper, the network domain security rules we set up are described as allowing data to pass through the source port address, the port on the destination address, and the service on the destination address. Digital domain security protection rules mainly describe the information to prevent unauthorized access. The main method is to encrypt the stored or transmitted information, that is, to decrypt the file requires a secret key or password.

\subsection{Problem Definition}
The problem of cyberspace attack and defense is a complex one, but it can be described as a game problem from the perspective of policy. In the model, both sides of the attack and defense are defined as the attacker and the defender, respectively representing the attacker and defender in multiple domain cyberspace. We divide the time slice of continuous into time slices, each time one and can only contain a cyberspace state, but the state is likely to be the same in different time slices. In each time slice, attacker and defender can choose an action to change the present state of cyberspace, the goal is to make the defender's total reward is maximum. When both the attacker and the defender select an action, the cyberspace state changes to the next state. The attacker and the defender check the current cyberspace state again, and then select an action based on their own policies to obtain rewards from the environment. The next cyberspace state is under the joint action of both attacker and defender. The purpose of proposed method in this paper is to make the defender get long-term, high and stable rewards in the multiple domain cyberspace attack and defense game.

Under completely rational conditions, both attacker and defender can estimate Nash equilibrium, which is generally considered to be the optimal strategy of both sides. But in practice, because of attacker and defender won't execute Nash equilibrium strategy from the start, but in the reinforcement learning method of their attack or defense strategy was improved, which means that reach the Nash equilibrium is not the result. Because of the different learning mechanism and strategy, could reach Nash equilibrium to deviate from the equilibrium.

The game theory-based model constructed precisely solves the problem game in this aspect, that is, defenders choose from the list of actions or strategies allowed to be selected at the same time or successively, one or more times under certain rules under certain environmental conditions, and then implement them and obtain corresponding results. From the point of view of game theory, we can think of attack and defense process as a game between attackers and defenders. Each side of the game chooses its own actions based on its own observation of the cyberspace environment and the prediction of the other side's actions. It can be considered that in the process of this game, the information of both sides is incomplete. As the game process progresses, both attacker and defender can obtain more information about the other side.

In the above game model, both the attacker and the defender are rational participants. For both sides, they have certain training data and know some attack policy or defense policy of the other side.

\subsection{Game model}

The game model is composed of each state of cyberspace attack and defense, the state transition between the actions and states chosen by attacker, defender in this state.

For each cyberspace state, we define the state as follows:

Constrained by bounded rationality, the benefits of attacker and defender from the cyberspace environment are set as their private information, unknown to the outside world, and the cyberspace state is set as the visible information of both sides, and the existing cyberspace state information can be observed by both sides.

Definition of time slice:

In general, each agent will observer cyberspace state, and then perform an action. We set the attacker and defender perform action at the same time, namely under the same condition, both attacker and defender choose an action and make cyberspace state transition to the next state. It does not exist under the same condition, attacker first action selection, following defender selects an action or the defender first selects an action and the attacker selects an action later.

Definition of cyberspace state transition model:

Probabilistic model is adopted to represent the randomness of cyberspace state transition. Markov property means that the current cyberspace state is only related to the previous cyberspace state, so our transition probability is, where, $S_t$ is the cyberspace state, and $(a_t, d_t)$ is the attack and defense actions of the attacker and the defender.

On the above basis, the relevant game model is constructed.

Definition 1: In cyberspace attack and defense, a standard form of game can be defined as: where $N=(attacker, defender)$ is the attacker and defender of the two agents participating in the game; $S = (s_1, s_2,... , s_n)$ is the set of game states, which refers to the state of the cyberspace in the attack and defense game. $A = (A_1, A_2,... , A_n)$ is the defense action set, where $A_k={a_1, a_2... A_m}$ is the action set of the defender in the network state $S_k$. $R(s_i, a_i, s_j)$ is the rewards the defender gets from the environment after executing the action $a_i$ in state $s_i$. $Q_d(s_i, a_i)$ is the value function of the defender's state action, indicating that the defender reaches a certain state $s_i$ after adopting a certain policy in the defense, and the expected rewards after executing the action $a_i$ in the state $s_i$ after continuing to execute the action according to the policy.

Definition 2: The cyberspace attack and defense of the Nash equilibrium is refers to in the game process, after the cyberspace game model both sides of the policies choices of several times, to get both sides unwilling or does not change his policy alone or policy combination, in this kind of attacker and defender both sides are reluctant to change their policy is called a Nash equilibrium policy situation.

The Nash equilibrium in cyberspace game model can be expressed as follows: Nash equilibrium is the optimal attack and defense policy of both attacker and defender in the game state $s_i$, and the attacker and defender can maximize their rewards only by adopting Nash equilibrium.

Theorem 1: Nash equilibrium exists. Given a game model, if the state set $S$ and attack and defense action set $A$ is a finite set in the game, then there must be a Nash equilibrium.

Definition 3: Defense effectiveness: For the attack action $a_i$, the defender also adopts defense action $d_i$. For the effectiveness of the defense action $d_i$, we use $\alpha(a_i,d_i)$ to represent it, that is, when the attack of attacker is failure, $\alpha(a_i,d_i)=1$; if the attack of attacker succeeds, the defense is judged to be failed, $\alpha(a_i,d_i)=0$. In general, $0<\alpha(a_i,d_i)<1$.

Definition 4: Defense cost ($DC$): For the defender, a series of defense actions needs related costs, such as the restart service, adding a firewall policy, etc. Also, defense actions may be limit other user's normal use, so we define the defense cost is sum of operating cost and negative defense costs.

Definition 5: Attack Reward ($AR$), represents the reward obtained by the attacker, is defined as

\begin{equation}
	AR=(1-\alpha)DC(a)+DC(d)-AC(a)
\end{equation}

Where, $DC(a)$ represents the system loss cost caused by the attack action $a$, $DC(d)$ is the defense cost required by the defense action $d$, and $AC(a)$ is the cost required by the attacker to launch the attack.

Definition 6: Defense Reward ($DR$), represents the reward the defender gets from the environment for taking an action.

\begin{equation}
	DR=AC(a)-(1-\alpha)DC(a)-DC(d)
\end{equation}

Where, $AC (a)$ represents the attack cost, the cost required by the attack action $a$, $DC(a)$ represents the cyberspace loss cost, and $DC(d)$ is the defense cost required by the defense action $d$.

Here, we set $AR+DR=0$, and we define multiple domain cyberspace game as a zero-sum game. In this model, the goals and returns of both sides are opposite. Although the rewards may be different, the benefits of both sides are complementary, so it is reasonable to adopt zero-sum game.

\subsection{Game Model based on Reinforcement Learning}

Attack and defense game model based on reinforcement learning is an intelligence technology, represented by deep reinforcement learning, which have been trained with an agent that can perceive and recognize the environment. With rules and learned experiential knowledge, defender pass through the attacker's actions, to realize the choice of optimal action. Game model policy is the mapping of state to action in game model, and it's the policy function $\pi$ in deep reinforcement learning.

\begin{figure*}[ht]
	\begin{centering}
		\centerline{\includegraphics[width=14cm]{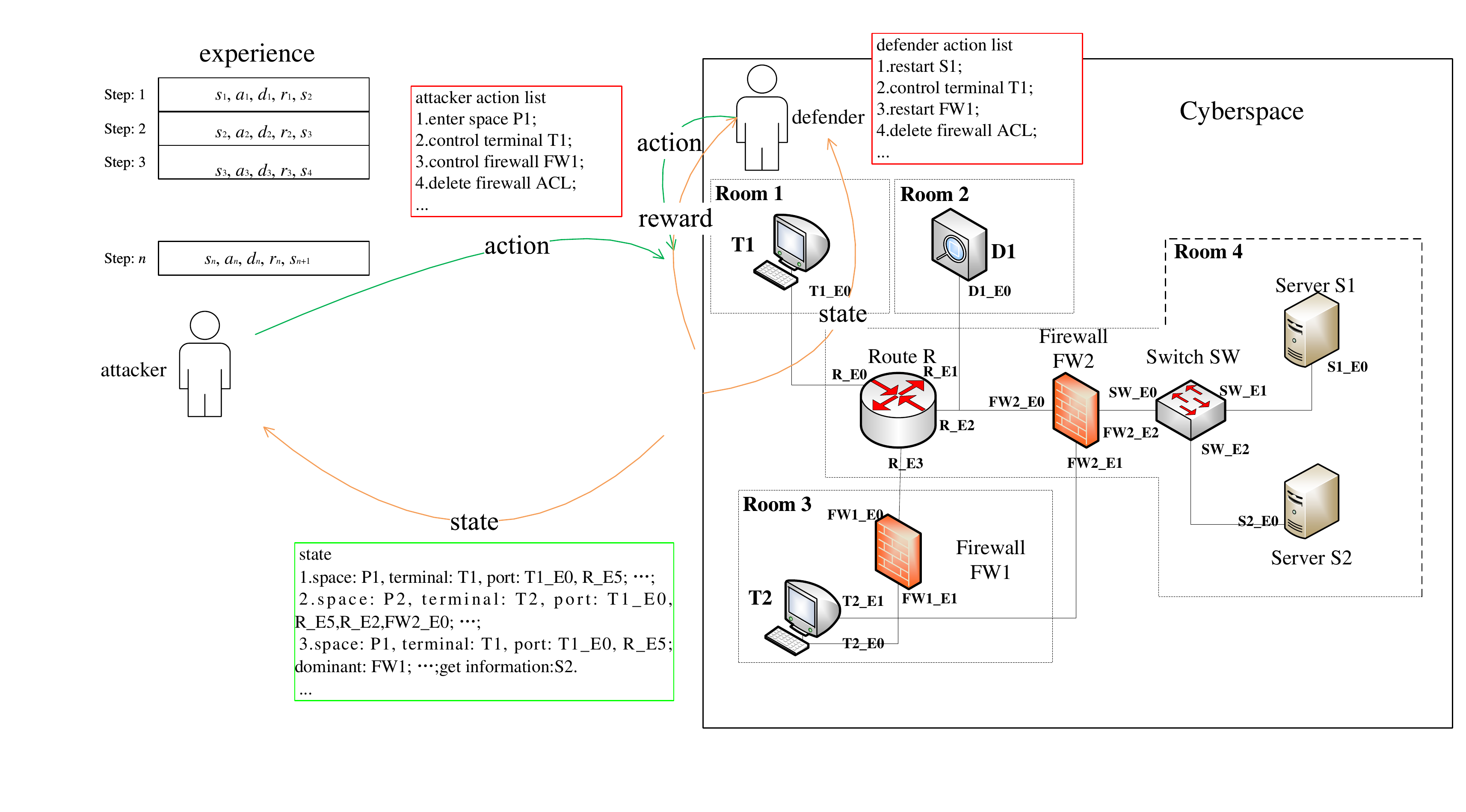}}
		\caption{Attack and Defense Game Model}
		\label{Figure1}
	\end{centering}
\end{figure*}

On the basis of describing the process of game based on MDP, this section constructs the game model based on deep reinforcement learning. Through the interaction between the agent based on reinforcement learning and the simulation cyberspace environment, the defender game policy under game task is trained and generated. The intelligent game defense model based on reinforcement learning method is shown in Fig. \ref{Figure1}.

Compared with the general single agent reinforcement learning, the single agent becomes the multiple agent. In this paper, we set one attacker and one defender here. On this basis, the defense algorithm is given, and we use DDPG as a defense algorithm for defender.

The algorithm first initializes the attack and defense game model and related parameters, defender agent will observer cyberspace states and perform defense actions from the algorithm. The next step is that the defender detects the cyberspace's state, and the next step is to make defense decision, select related defense actions according to the current policy. Then it's time to update the algorithm's parameters based on the reward, with the goal of maximizing the defender's reward. Every time the agent chooses an action, it will adjust its learned policy according to the action feedback and the new cyberspace state, and the cycle repeats. After constant trial and error and continuous interaction with the environment, the agent finally learns the optimal policy of game in the cyberspace environment.

\subsection{Reward Randomization Reinforcement Learning}

In order to improve the learning efficiency of reinforcement learning and reduce the dependence of reinforcement learning on data, we introduce the reward randomization method to set the reward of defender agent, because in reinforcement learning training, agent's reward function is difficult to set. Because in reinforcement learning, the reward function $R$ highly influences the polices that reinforcement learning may learn. Further, even if a policy is difficult to learn in a particular $R$, it may be easier to $R_0$ other functions in some $R$. Therefore, if we can define a suitable space $R$ on different state and draw samples from $R$, we may discover the ideal new policy to run reinforcement learning on some sample reward function $R_0$ and evaluate the effect of the obtained policy on the original game. 

\begin{figure}[h]
	\begin{centering}
		\centerline{\includegraphics[width=7cm]{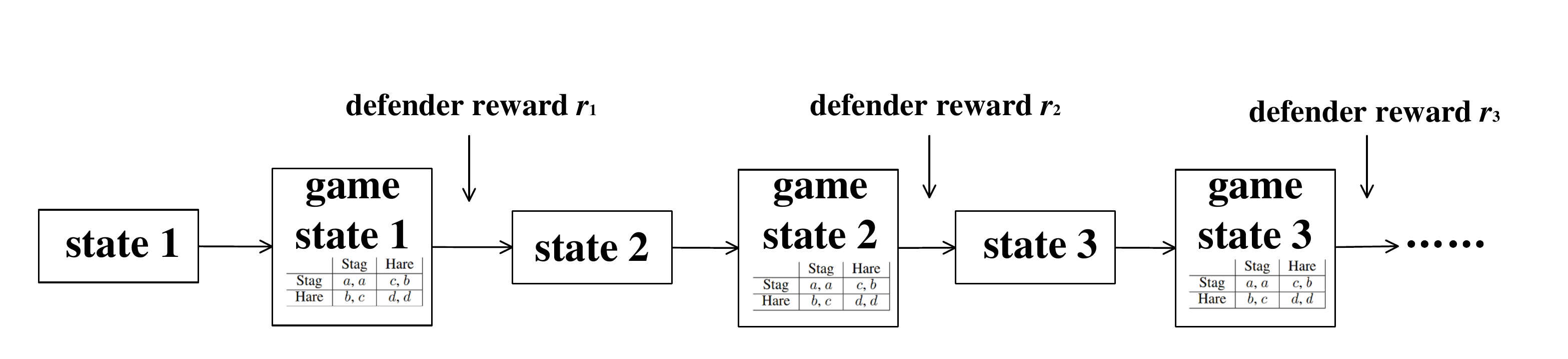}}
		\caption{Game Model Process}
		\label{Figure2}
	\end{centering}
\end{figure}

In this paper, we extend reward randomization to general multiple agent game. Considering the current game model is defined as follows: $S$ is the observable space state, $A$ is the action space of the defender, and the defender agent generates the policy $\pi(\theta_a)$, which is the parameter of the policy. In this game model, both the attacker and the defender hope to get the maximum reward. The defense game model process is shown in Fig. \ref{Figure2}.

In general, for Nash equilibrium, the desired outcome is to maximize the rewards for both sides. But when it comes to cyberspace attack and defense, we only expect the defender to maximize his reward. Therefore, this problem can be equivalent to: Under the above conditions, we need to find a policy that maximizes defender's rewards. Because the two sides of the attack and defense is a zero sum game, that is the attacker gains the least.

The existing problem is that although the current decentralized policy method is a popular learning method for complex Markov games, the strategies it eventually trains are often suboptimal, that is, there will be a policy. We still can't find this with our current methods.

Therefore, reward randomization is introduced into the attack and defense game model is based on the following reasons: if the relevant reward function in the game makes it difficult for reinforcement learning to discover optimal policies, it may be easier to achieve this goal if the reward function is interfered with. Therefore, we can define a reward function space $R$, so that we can train from the reward function sampling and calculate the associated optimal policy, instead of simply learning a fixed reward. Formally, we define a subspace of the reward function to guide the training of reinforcement learning in the game. By using the reward randomization method, we take $n$ samples and get corresponding policies in each training. Get the corresponding policies by calculating the original game. Finally, we can further improve the performance of the model by fine-tuning the policy. This training process is called reward randomization reinforcement learning. We define the linear mapping reward randomization as follows:

\begin{equation}
	R_t=
	\left  \{
	\begin{array}{l}
		R_t \quad if \; random\ge0.5\\\\
		random(R_n) \quad if \; random<0.5
	\end{array}
	\right.
\end{equation}

which $random$ is a randomly generated data in experiments and $0\le random\le 1$, $random(R_n)$ means taking a sample from reward space $R_n$.

In general, feature-based reward functions are often found in relevant reinforcement learning reference. For example, in games with relevant distance, reward is usually set as the negative distance from the target position to the agent's position plus the most successful reward. In the general reinforcement learning game, it can be a single reward for each agent, or the reward of the whole group, or it can be set up to give more emphasis to the agent.

Finally, we fine-tune the reward. Generally, fine-tuning has two benefits: Firstly, the policy found in the perturbation game may not maintain equilibrium in the original game, and fine-tuning can ensure the final results convergence; Secondly, in experiments, fine-tuning can further help to get rid of suboptimal policy.

\section{Experiments}

\subsection{Experiment Cyberspace Environment}
\begin{figure}[ht]
	\begin{centering}
		\centerline{\includegraphics[width=7cm]{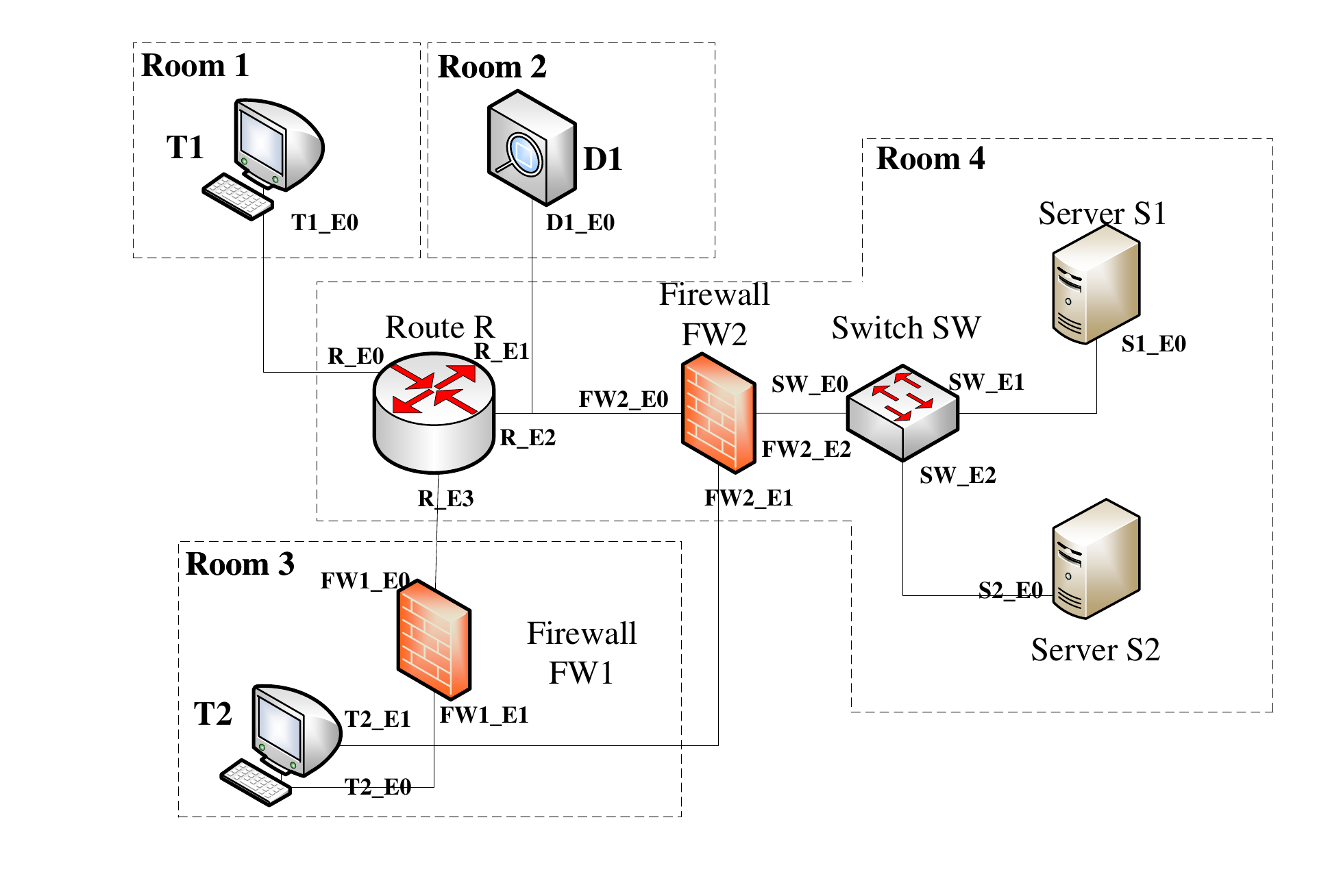}}
		\caption{Experiments Cyberspace Environment Topology}
		\label{Figure3}
	\end{centering}
\end{figure}

We used Python to simulate a cyberspace environment in this experiment. There are a total of five places in this experimental environment. A physical space, such as a district, a college, is represented by the outermost layer. Room 1 is the terminal's location, Room 2 is the security equipment's location, Room 3 is the terminal's location, and Room 4 is the service's location. As shown in Fig. \ref{Figure3}, there are five types of equipment: computer (T1 and T2, respectively stored in Rooms 1 and 3), firewall (FW1 and FW2, respectively stored in Rooms 3 and 4), security equipment (D1, stored in Room 2), router (R, stored in Room 4) and switch (SW, stored in Room 4), server (S1, S2, stored in Room 4) and their equipment connection relationships. S2 is where the security information is kept. If an attacker is able to gain access to S2 in order to obtain the security file, the attack has been successful.

\subsection{Baselines}
\begin{figure*}[ht]
	\centering
	\subfigure{
		\centering
		\includegraphics[width=3.2in]{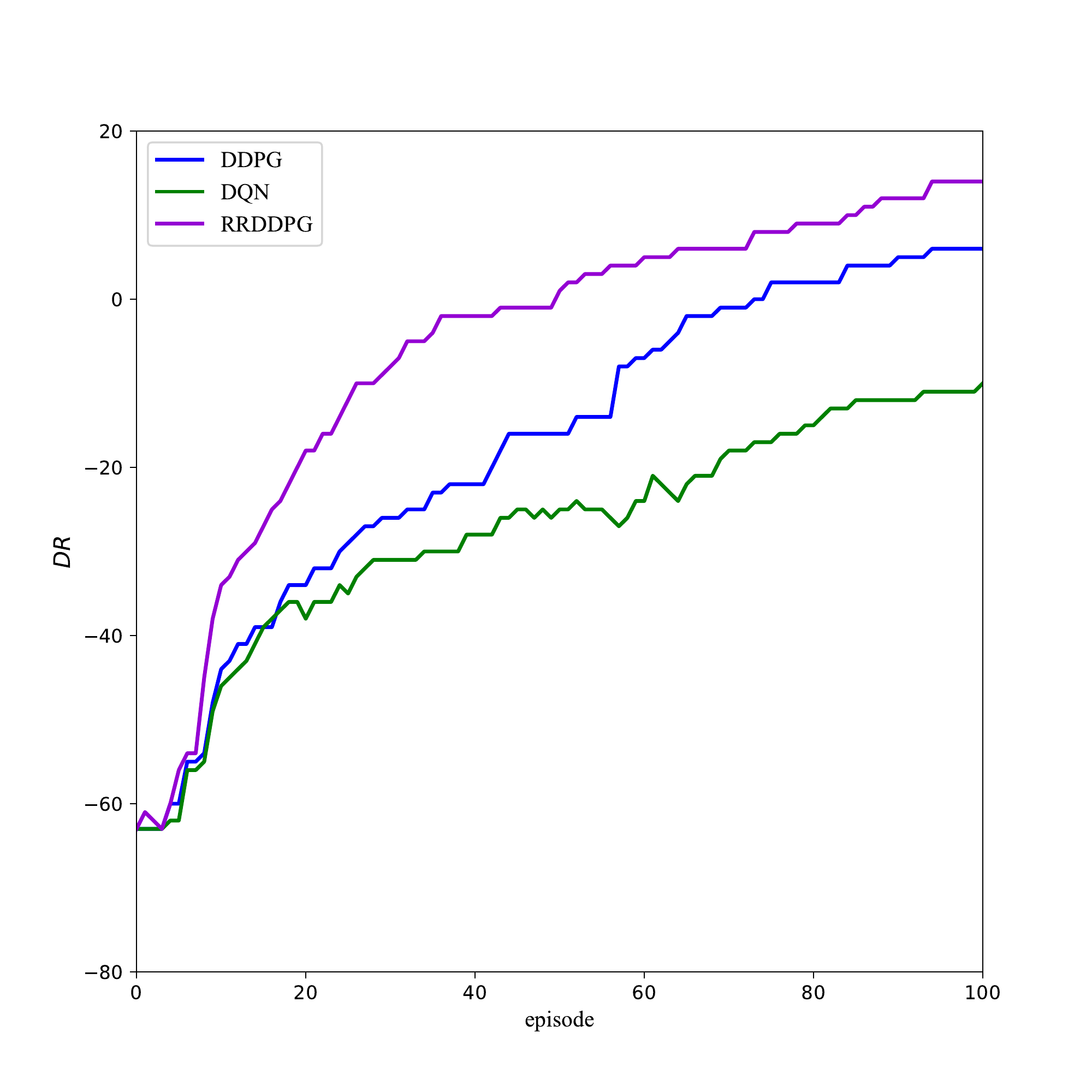}
	}	
	\subfigure{
		\centering
		\includegraphics[width=3.2in]{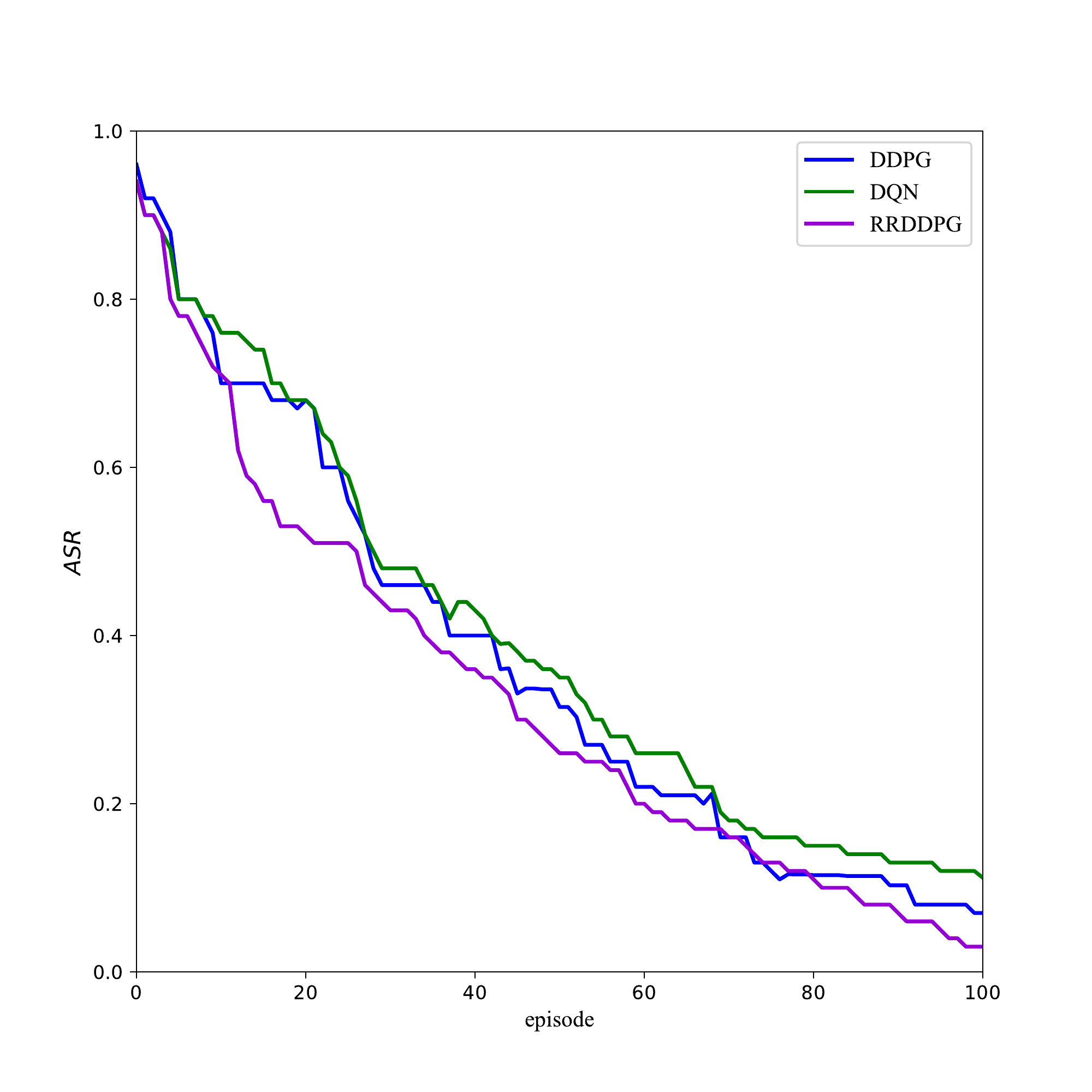}
	}	
	\caption{$ASR$ and $DR$ in Different Methods}
	\label{Figure4}
\end{figure*}

Experiments are carried out in the above-mentioned experimental context. The following baseline approaches are used to demonstrate the effectiveness of our proposed method:
\begin{enumerate}
	\item Deep Q Network (DQN): DQN is a typical reinforcement learning that employs a neural network to predict Q-value and constantly updates the neural network to learn the best action path. In DQN, there are two neural networks. The target network, which is relatively fixed, is used to obtain the value of Q-target, while the evaluate-network is utilized to obtain the value of Q-evaluate.
	\item DDPG. This method is described in the section "Related Works."
\end{enumerate}

Attackers are guided by knowledge-based rules, and their attack goal is to successfully attack the server. The following are the specific attack steps:

First, attacker enters space Room 2 and obtains the firewall FW1's management service password, FW1 password;

Second, attacker uses device T1 or D1 to gain access to FW1's management service; add the following access control list: allow T1 or D1 to gain access to T2's management service, T2 manager;

Third, attacker uses T2 manager to obtain the password FW2 password for the firewall FW2 and the password S1 web password for the service S1 web;

Fourth, attacker opens the T2 S1 port, log in to the FW2 manager firewall, and adds the following access control list: allow T1 or D1 access to the S1 web and S2 web services;

Fifth, attacker visits the service S1 web using T1 or D1 and obtain the password S2 web password of S2 web;

Finally, the attacker can utilize T1 or D1 to get access to the S2 web service and obtain the security file using the S2 web password. At this point, the attacker is finished his attack.

\subsection{Results and Discussion}

For each method, we train 100 episodes. In each episode, 100 attackers were generated at random for training purposes in order to assess the method's effectiveness. After 100 episodes, the DQN, DDPG, and our suggested reward randomization DDPG method were tested for performance. Meanwhile, we've selected the Assault Success Rate (ASR) as the criterion for evaluation. We define following:

\begin{equation}
	ASR=\frac{attack\ successfully \;attackers}{total\; number\; of\; attackers }
\end{equation}

The total number of attackers in this experiment is 100 in an episode.

Each time slices the attacker performs an action in the experiment, the attacker performs an action and the defender perform an actions and receives reward. Rewards for all agent actions were added as a training reward during the model training phase. During each training session, the model will be trained 100 times. The following are the additional parameters in the experiment: The attacker's ratio in the experimental cyberspace environment is \emph{UP}=0.4, the attacker's ratio of launching an attack is \emph{AP}=0.3, and each user performs no more than 60 actions. If attacker cannot complete the attack in 60 time slices, he will be log down this cyberspace environment. Also, if attacker attack successfully, he will get $AR=10$, defender will get $DR=-10$; if not, he will get $AR=-10$, defender will get $DR=10$. Also, if attacker have no harvest in 60 time slices and he will not captured by defender, he will get $AR=-5$, and defender will get $DR=0$.

Firstly, the game model was realized in the above experimental cyberspace environment. At the end of each episode of method training, we recorded the $ASR$. The experimental results are shown in the figure Fig. \ref{Figure4}.

Secondly, in order to evaluate the performance of RRDDPG method, we recorded $DR$ in different methods, and the experimental results are shown in Fig. \ref{Figure4}.

From the experimental results, with the increase of training times, the $ASR$ is constantly decreasing, while the defender's $DR$ is constantly increasing, which indicates that the multiple domain cyberspace attack and defense game based on reinforcement learning can effectively improve the defense ability of defenders. At the same time, the method we proposed makes the attacker have the lowest $ASR$ and the highest $DR$, which shows the advanced performance of the our proposed reward randomization DDPG method.

Meanwhile, in the multiple domain cyberspace attack and defense game, from the experimental results, the following conclusions are:
\begin{enumerate}
	\item{The multiple domain cyberspace attack and defense game system designed in this paper can provide a research direction for the next step of cyberspace attack and defense game intellectualization;}
	\item{A reward randomization DDPG method is proposed, it can improve the intelligent defense level of defenders in multiple domain cyberspace attack and defense games;}
	\item{In this model, we assess the defense and attack rewards and include them into the model such that the defensive option becomes more effective and more in line with real-world applications.}
\end{enumerate}

\section{Conclusion}
Due to the complexity of the offense and defense game, the optimal defensive policy selection problem in multiple domain cyberspace is a challenge task. The description process of the cyberspace attack and defense game is incompatible with the present multiple domain cyberspace security model. Both attack and defensive plans in a multiple domain cyberspace environment are chaotic, and they can't effectively reflect the attacker's and defender's actions and state changes, as well as the game process. Therefore, we introduced the multiple domain cyberspace include the physical domain, the digital domain and the network domain. The model of the attacker and defender can be observed by both attacker and defender to the state to determine the next step of game action, for the defense in a number of defensive states dynamic search for the optimal defensive policy, by building a game model of cyberspace attack and defense. Following, we establish the multiple domain cyberspace attack and defense of the dynamic game model. Simultaneously, we consider cyberspace attack and defense as a zero-sum game, and uses this to propose the reward randomization DDPG algorithm to maximize the defense's rewards. Experimental results reveal that the proposed method have superior defense effects.


%



\ifCLASSOPTIONcompsoc
  \section*{Acknowledgments}
\else
  \section*{Acknowledgment}
\fi

Throughout this manuscript, I have received a great deal of support and assistance.

I would like to thank my supervisor, Xingchun Diao, for his guidance through each stage of the process. 

I would like to acknowledge Professor, Zhisong Pan, for inspiring my interest in the development of artificial intelligence innovative technologies. 

My research partner, Yu Pan, for her wise counsel and sympathetic ear of my research, you are always there for me. For this, I am extremely grateful.

In addition, I would like to thank the support of the National Natural Science Foundation of China No.62076251 and No.62106281.

\ifCLASSOPTIONcaptionsoff
  \newpage
\fi



\bibliographystyle{IEEEtran}
\bibliography{IEEEabrv}
%



%

\begin{IEEEbiography}[{\includegraphics[width=1in,height=1.25in,clip,keepaspectratio]{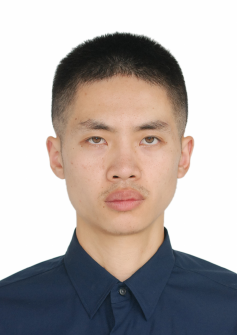}}]  
{Lei Zhang} was born in 1989 and received master's degree in Software Engineering from Army Engineering University of PLA, Nanjing, in 2018. He is currently pursuing the Ph.D. in Computer Science and Technology in Army Engineering University of PLA, Nanjing, China, 210007. His main research interests include reinforcement learning, machine learning, data mining and network security. \\
E-mail:zhanglei@aeu.edu.cn.
\end{IEEEbiography}
\begin{IEEEbiography}[{\includegraphics[width=1in,height=1.25in,clip,keepaspectratio]{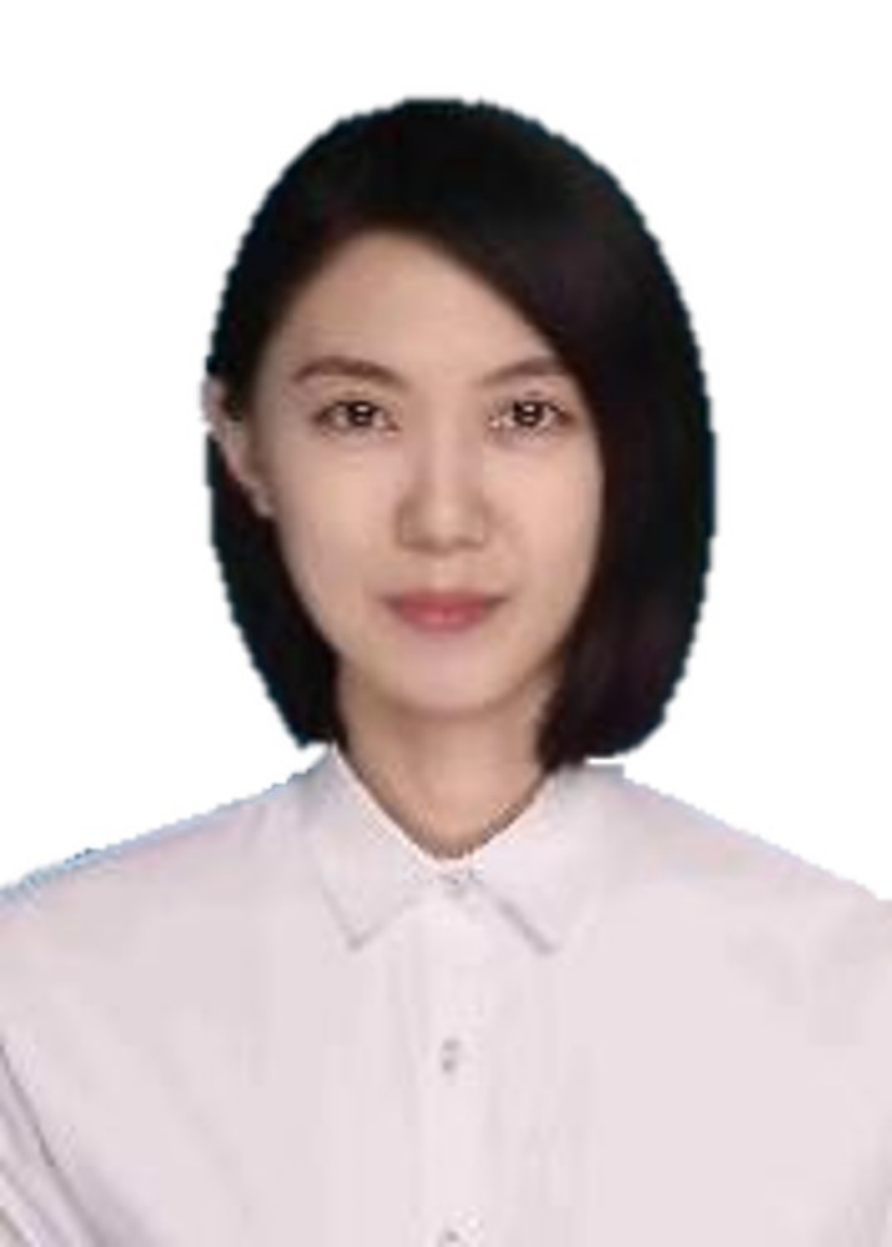}}] 
{Yu Pan} was born in 1990 and received master's degree in Computer Science and Technology from Northeastern University, Shenyang, in 2015. She received the Ph.D. in Computer Science and Technology from Army Engineering University of PLA, Nanjing, in 2021. She is now an engineer in 31436 troop of PLA, Shenyang, Liaoning, 110005. Her main research interests include machine learning, data processing and mining in social networks.  \\
E-mail: panyu@aeu.edu.cn.
\end{IEEEbiography}
\begin{IEEEbiography}[{\includegraphics[width=1in,height=1.25in,clip,keepaspectratio]{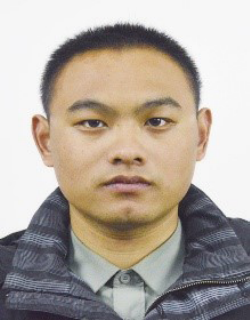}}] 
{Yi Liu} was born in 1990 and received master's degree in Computer Science and Technology from the PLA University of Science and Technology, Nanjing, in 2014. He received the Ph.D. in Software Engineering from Army Engineering University of PLA, Nanjing, in 2018. He is now an assistant researcher in Academy of Military Science, Beijing, China, 100091. His main research interests include machine learning, evolutionary algorithms, and data quality. \\
E-mail: albertliu20th@163.com.
\end{IEEEbiography}
\begin{IEEEbiography}[{\includegraphics[width=1in,height=1.25in,clip,keepaspectratio]{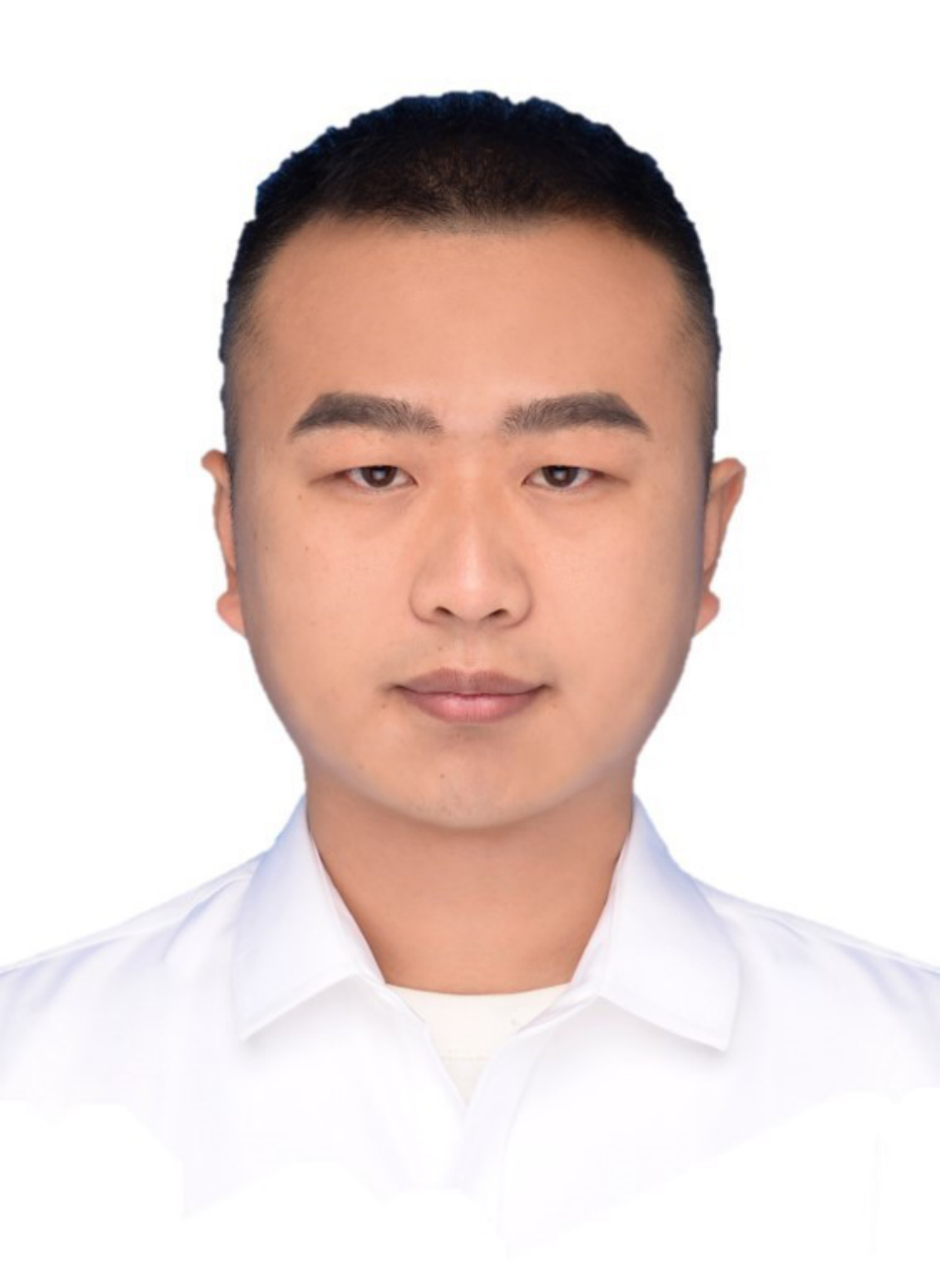}}] 
{Qibin Zheng} was born in 1990 and received master's degree in Army Engineering University of PLA, Nanjing, in 2016. He received the Ph.D. in Software Engineering from Army Engineering University of PLA, Nanjing, in 2020. He is now an assistant researcher in Academy of Military Science, Beijing, China, 100091. His main research interests include data mining, machine learning and multiple modal data analysis.  \\
E-mail:zqb1990@hotmial.com.
\end{IEEEbiography}
\begin{IEEEbiography}[{\includegraphics[width=1in,height=1.25in,clip,keepaspectratio]{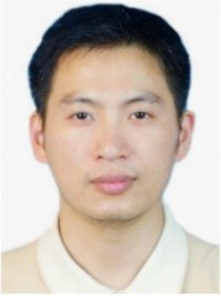}}] 
{Zhisong Pan} was born in 1973 and received Ph.D. from the Department of Computer Science and Engineering, Nanjing University of Aeronautics and Astronautics in 2003, Nanjing, China. Since July 2011, he has worked as a full professor and Ph.D. supervisor at Command and Control Engineering College, Army Engineering University of PLA, Nanjing, China. His current research interests include pattern recognition, machine learning and neural networks. 
\\ E-mail: panzhisong@aeu.edu.cn.
\end{IEEEbiography}







\end{document}